\documentclass[letterpaper, 10 pt, conference]{ieeeconf}  

\IEEEoverridecommandlockouts                              

\usepackage{graphicx}
\usepackage{algorithm,algorithmic}
\usepackage{amsmath}
\usepackage{subcaption}
\usepackage{multirow}
\usepackage[table]{xcolor}

\newcommand*{\prob}{\mathsf{P}}

\title{\LARGE \bf
Robust and Fast 3D Scan Alignment using Mutual Information
}

\author{Nikhil Mehta$^{1}$, James R. McBride$^{2}$ and Gaurav Pandey$^{2}$
\thanks{*This work was supported by Ford Motor Company}
\thanks{$^{1}$N. Mehta is with the Computer Engineering Department of IIT Kanpur, India. {\tt\small nikhil@cse.iitk.ac.in}
}
\thanks{$^{2}$G. Pandey and J. R. Mcbride are with Ford Motor Company,
Dearborn, USA. {\tt\small gpandey2@ford.com},  {\tt\small jmcbride@ford.com.}
}
}

\begin{document}

\maketitle
\thispagestyle{empty}
\pagestyle{empty}

\begin{abstract}

This paper presents a mutual information (MI) based algorithm for the estimation of full 6-degree-of-freedom (DOF) rigid body transformation between two overlapping point clouds. We first divide the scene into a 3D voxel grid and define simple to compute features for each voxel in the scan. The two scans that need to be aligned are considered as a collection of these features and the MI between these voxelized features is maximized to obtain the correct alignment of scans. We have implemented our method with various simple point cloud features (such as number of points in voxel, variance of z-height in voxel) and compared the performance of the proposed method with existing point-to-point and point-to-distribution registration methods. We show that our approach has an efficient and fast parallel implementation on GPU, and evaluate the robustness and speed of the proposed algorithm on two real-world datasets which have variety of dynamic scenes from different environments.


\end{abstract}

\section{INTRODUCTION}
A 3D alignment algorithm to determine the relative rigid body transformation between two partially overlapping scans is an underpinning tool for many applications in mobile robotics including localization, mapping and navigation systems. In this work, we consider a robot which obtains two 3D scans ($A$ and $B$) from two poses $P_0$ and $P_1$ via a 3D laser scanner. Provided that some part of the environment is common to both scans, it is generally possible to find a rigid-body transformation $T$ that can project the points in $P_1$ so that they align with $P_0$. The solution to the process of scan alignment ($T$) is parameterized by six values: three translation components ($t_x$, $t_y$ and $t_z$) and three rotation components ($\theta_x$, $\theta_y$ and $\theta_z$). The reason that scan alignment problem is at the center of most navigation, mapping and localization systems, is simply because the rigid body transformation $T$ derived from alignment is of higher quality than odometry estimate (due to wheel slippage and surface irregularities).

The primary challenge in the problem is to minimize the runtime complexity while maximizing the robustness of the solution. Most existing methods are either designed around computationally-efficient local searches which are not robust to initialization or global search methods which are computationally intense. At vehicle's dead-reckoning error, the initial estimate can be far from global maximum resulting in large errors when using local-search methods. To counter this problem, most implementations currently involve running the registration algorithm at high frequency rate resulting in a constant load on computation resources. 

\begin{figure}[t]
    \centering
    \includegraphics[width=0.95\linewidth]{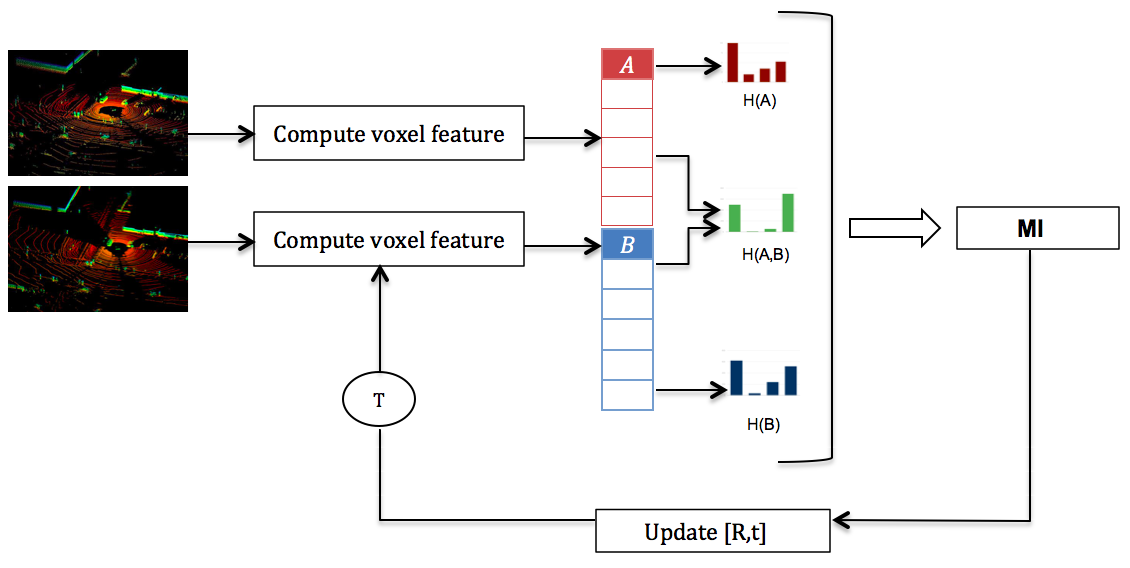}    
    \caption{An overview of mutual information (MI) based scan alignment framework. MI maximization is applied over the voxelized-features computed from two partially overlapping scans. }
    \label{fig:algo_flow}
\end{figure}

Our approach poses the alignment problem in a MI (mutual information) maximization framework: it finds the rigid body transformation that maximizes MI between the features of two scans. An overview of our approach is illustrated in Figure \ref{fig:algo_flow}. We show that even primitive features like variance and number of points give results better than the other state of the art point-to-point and point-to-distribution methods for large prior errors from odometry. 

The central contributions of this paper are:
\begin{itemize}
\item We present a robust and fast framework for scan alignment by posing the problem in a MI maximization framework with simple voxelized features. 
\item Unlike other 3D scan alignment methods, our approach allows us to consider the no-feature voxels (unoccupied region in scene).
\item We present a detailed empirical evaluation of our method in different environments: both urban and rural scenes. We compare our approach with a point-to-point and a point-to-distribution alignment method.
\item We show how the proposed approach has an efficient fast and robust GPU implementation, freeing the CPU for other important tasks.
\end{itemize}

The quality and robustness of our method along with it's ability to work in real-time, makes it ideal for mobile robotic systems in which accuracy is of high importance. 

In the following section we present a brief overview of the prior work (Section \ref{related_work}). In section \ref{approach}, we describe the proposed framework. Empirical evaluation and comparison with other methods along with runtime analysis is shown in section \ref{evaluation}.


\section{RELATED WORK} \label{related_work}
Iterative Closest Point (ICP) \cite{Besl:1992:MRS:132013.132022} is one of the most popular scan registration algorithms used to estimate the optimal transformation between two overlapping scans. In ICP, closest points between scan $A$ and reference scan $B$ are used to obtain a closed-form solution by optimizing the sum of squared distances (usually Euclidean distance). Performing the nearest neighbor search in ICP becomes a bottleneck due to its high computational cost, however, using a K-D Tree \cite{1240280} does mitigate the problem to an extent. Iterative Dual Correspondence (IDC) \cite{Lu1997} generates corresponding points for both rotation and translation separately, with optimization done in an alternate fashion. This improves the alignment accuracy when the initial estimate has large rotational error. Iterative Closest Line (ICL) \cite{55667789-MIT}, \cite{297537764-MIT}, \cite{Censi08anicp} is a variant of ICP in which instead of matching the points in both the scans, the query points in scan $B$ are matched to lines extracted from points in reference scan $A$. Generalized ICP (GICP) \cite{Segal-RSS-09}, attaches a probabilistic model in estimating the correct corresponding points by taking the covariance structure derived from the local neighborhood in the environment. 

Another widely used algorithm is Normal Distribution Transform (NDT) which was initially proposed for 2D data \cite{Biber2003TheND} and was later extended to three-dimensions. 3D-NDT \cite{Magnusson053dmodelling} is a point-to-distribution method in which the maximum likelihood estimate (MLE) of points in $B$ is maximized over the distribution of points in a gaussian mixture model derived from $A$. Supervoxel based NDT (SV-NDT) \cite{kim2016} is an extension to 3D-NDT in which segmentation on the basis of local-spatial structure is done over $A$ before creating a GMM model. 

The problem of scan alignment with fused sensor data as input has also been extensively researched in the past two decades. Some of the work includes incorporating color information in the ICP framework (\cite{Johnson:1997:RIT:523428.825385}, \cite{Godin2001AMF}) by augmenting RGB color channels to 3D coordinates and by exploiting the co-registration of 3D data with the available camera imagery to associate scale invariant feature transform (SIFT) \cite{Lowe:2004:DIF:993451.996342} or speeded up robust features (SURF) \cite{Lowe:2004:DIF:993451.996342} features to the 3D points as in \cite{f8a4caebc7cf412ca3ed2113bf6c94df}. 

Mutual information based alignment methods was first proposed in \cite{pmid8331222} for multi-modality medical images. Since then literature has been filled with work inspired by mutual information, these include minimization of distribution of joint histograms \cite{Hill1993}, data alignment from multiple modalities (different sensors) as in \cite{Viola1997}, \cite{pmid9101328}, \cite{pmid10709702}, and \cite{6386053}. In \cite{1216223}, a comprehensive survey of mutual information based techniques being used in medical images is presented.

The proposed method is closely related to FPFH \cite{6386053} mainly because we use mutual information to compute the registration parameters. However, the feature selection and histogram creation method is different. In \cite{6386053} high dimensional features are computed using FPFH which are then quantized into one of the precomputed codewords. There is no denying that FPFH are better descriptors of a scan than simple voxelized features like z-variance, but in a MI based framework, we don't need to find correspondences between patches, therefore computing FPFH only increases the run-time complexity due to nearest neighbor search and normal computation at each point. Moreover, unlike voxelized features, FPFH fails to take into account the empty region in the scan.      

\begin{figure}[t]
\centering
\begin{subfigure}{.48\linewidth}
	\centering
    \includegraphics[width=\linewidth, height=\linewidth]{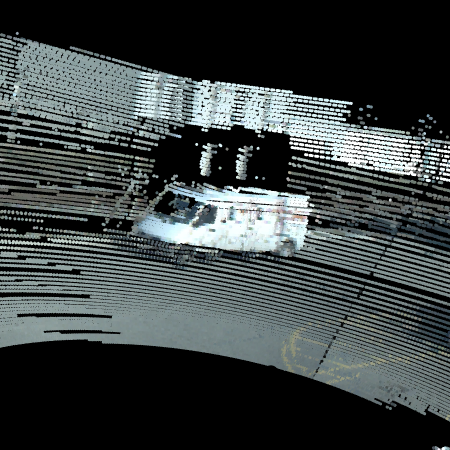}
    \caption{Unaligned scans}\label{fig:trans_base}
\end{subfigure}
\hspace{1mm}
\begin{subfigure}{.48\linewidth}
    \centering
    \includegraphics[width=\linewidth, height=\linewidth]{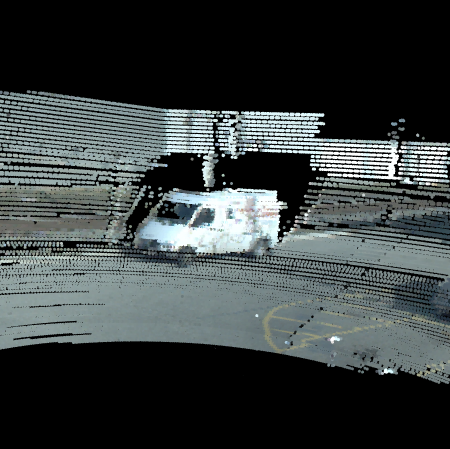}
    \caption{Aligned scans}\label{fig:trans_mi}
\end{subfigure}
\caption{This figure shows the cropped 3D textured point cloud of a pair of scan transformed to the same frame of reference. (a) Transformation used is from the initial estimate. (b) Transformation applied is first improved/aligned using our approach.} 
\label{fig:qualitative_analysis}
\end{figure}

\section{METHODOLOGY} \label{approach}

\subsection{Approach}
A qualitative result of the proposed MI based approach using voxelized features is shown in Figure \ref{fig:qualitative_analysis}. In order to estimate the correct rigid body transformation $T$, represented by a \([4 \times 4]\) transformation matrix, we first initialize our scene with a large 3D grid with resolution $R_v$ ($1\times1\times1$) that encompasses both the scans $A$ and $B$. We create two voxel-point mapping tables by finding the corresponding voxel (the voxel which contains the point) for all points in scan A and B. Now we calculate simple features like (\romannumeral 1) Variance of z-height, or (\romannumeral 2) number of points in voxel. It is important to note that unoccupied voxels are also included in our voxel-point mapping table with $\phi$ (no-feature) value; this results in a voxel-feature mapping that contains large number of voxels with feature $\phi$. Intuition behind this approach is that in a perfectly aligned state not only the occupied voxels should be aligned, but unoccupied voxels should also be aligned in the overlapping region.

To obtain the statistical dependence using mutual information of two partially overlapping scans, we define an overlapping region which is a subgrid parameterized by six variables $[x_{min}^o, x_{max}^o, y_{min}^o, y_{max}^o, z_{min}^o, z_{max}^o]$ (where 'o' is the overlapping region). We consider the feature maps of this overlapping subgrid as two random variables representing the structure of the environment. These random variables are used to calculate the mutual information of both the scans. The mutual information between these two random variables is the amount of information obtained about one variable, through the information about the other variable. Therefore, under the correct rigid body transformation, we can expect the information from one of the scans to give maximum amount of information about the other scan, thus rendering the MI to be maximum between the feature maps.

\begin{figure}[!t]
	\includegraphics[width=0.5\textwidth]{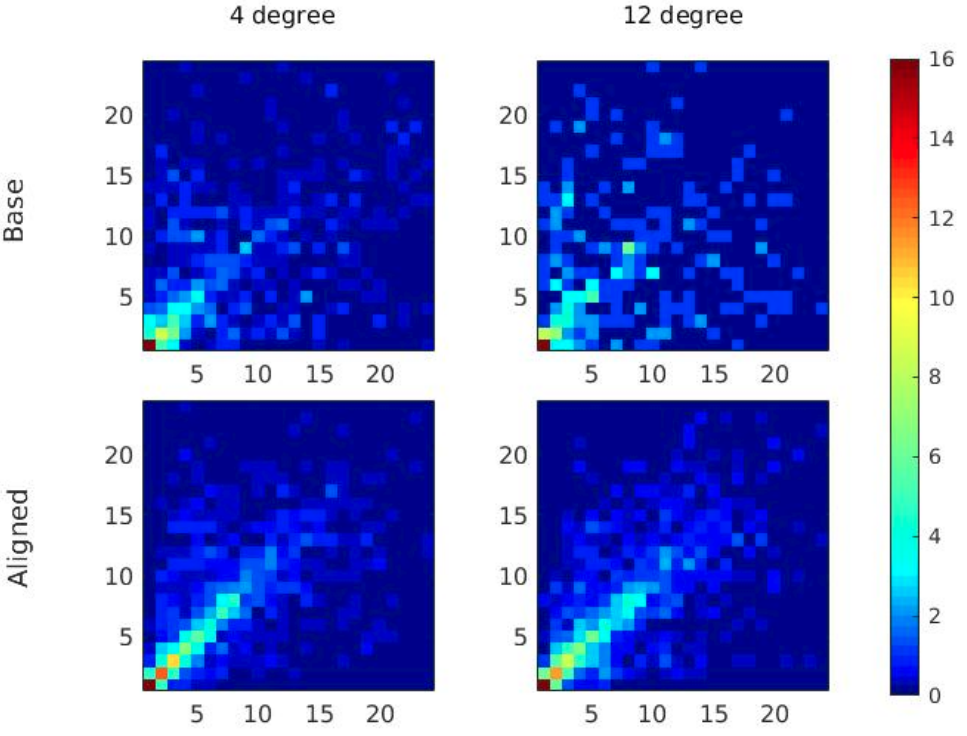} 
     \caption{This figure shows 4 joint histograms: 2 before alignment (Row 1) and 2 after alignment using the proposed method (Row 2). Each column represents two different test cases: in Column 1 the scan pair had the initial heading error of 4$^{\circ}$ and in Column 2 the scan pair had the initial heading error of 12${^\circ}$. X-axis and Y-axis of all the 4 joint histograms represent the variance of Z in a voxel for scan A and scan B respectively. Note how the dispersion in the joint histogram decreases (forming a straight line) after alignment in both the test cases. It can also be seen that the dispersion in joint histogram increases as we increase the heading error from Column 1 - Column 2 (4$^{\circ}$ to 12${^\circ}$). Best viewed in color.} \label{fig:joint_histos}    
\end{figure}

\begin{equation} \label{eq:cost}
\eta =  \operatorname*{arg\,max}_\eta MI(X, Y; \eta) 
\end{equation}

where \( \eta = [t_x, t_y, t_z, \theta_x, \theta_y, \theta_z] \) are the required transformation parameters, \( [X, Y] \) are the respective random variables representing the features of scan A and B in the corresponding overlapping region and $MI(X, Y; \eta)$ is the mutual information of the overlapping voxels under the transformation $\eta$.

\begin{equation}\label{eq:mi}
MI(X,Y; \eta) = H(X;\eta) + H(Y;\eta) - H(X,Y;\eta)
\end{equation}

where \( H(X;\eta) \) and \( H(Y;\eta) \) are the entropies of the random variables X and Y respectively, and \( H(X,Y;\eta) \) is the joint entropy of the two variables at transformation $\eta$.

\begin{equation}\label{eq:marginal_A}
H(X;\eta) = -\sum_{x \in X} \prob_X(x;\eta)\log {\prob_X(x;\eta)}
\end{equation}

\begin{equation}\label{eq:marginal_B}
H(Y;\eta) = -\sum_{y \in Y} \prob_Y(y;\eta)\log {\prob_Y(y;\eta)}
\end{equation}

\begin{equation}\label{eq:joint_AB}
H(X,Y;\eta) = -\sum_{x \in X}\sum_{y \in Y} \prob_{XY}(x,y;\eta)\log {\prob_{XY}(x,y;\eta)}
\end{equation}

Here $\prob_X(x;\eta)$ is the marginal probability for $X=x$, $\prob_Y(y;\eta)$ is the marginal probability for $Y=y$ and $\prob_{XY}(x,y;\eta)$ is the joint probability for $X=x$ and $Y=y$ when transformation parameters are $\eta$. For simplicity, we have taken x and y to be either the variance in z or the number of points in a voxel.

In each iteration, the MI value changes and is maximized when all voxels (occupied and unoccupied) are aligned. It is important to note that with every iteration, the overlapping region will change and the correct subgrid will be found when the MI is maximum. Figure \ref{fig:joint_histos} represents the joint-distribution before (row 1) and after (row 2) MI maximization for two pairs of scan with different initial error. It can be seen that the joint distribution converges to a line passing through origin and becomes relatively uniform after the correct alignment.

\begin{algorithm}[!t]
\caption{Alignment of two scans using MI framework}
\label{alg:alignment}
\begin{algorithmic}[1]
\renewcommand{\algorithmicrequire}{\textbf{Input:}}
\renewcommand{\algorithmicensure}{\textbf{Output:}}
 
\REQUIRE Two scans A and B with some partial overlap. Initial guess of the rigid-body transformation $T_o$.
\ENSURE  Estimated registration parameter T
\\ \textit{Initialization} :
\STATE Extract the associated voxel ids for all points in scan A
\STATE Create scan A feature map for all voxels
\\ \textit{Alignment} : 
\STATE $T_c = T_o$
\WHILE{MI not converged}
\STATE Transform scan B using current rigid body transformation $T_c$ 
\STATE Extract associated voxel ids for points in scan B
\STATE Create scan B feature map for all voxels
\STATE Define an overlapping bounding region: \label{alignment:line:8}

$x_{min}^o = max(x_{min}^A, x_{min}^B)$  

$y_{min}^o = max(y_{min}^A, y_{min}^B)$

$z_{min}^o = max(z_{min}^A, z_{min}^B)$ 

$x_{max}^o = min(x_{max}^A, x_{max}^B)$ 

$y_{max}^o = min(y_{max}^A, y_{max}^B)$ 

$z_{max}^o = min(z_{max}^A, z_{max}^B)$ 
\STATE Consider random variable X and Y as the voxelized feature map for all voxels lying inside the overlap of scan A and scan B respectively
\STATE Compute the marginal and joint entropy using the equations \ref{eq:marginal_A}, \ref{eq:marginal_B} and \ref{eq:joint_AB}
\STATE Calculate MI using \ref{eq:mi}

\STATE Update $T_c$ using Nelder-Mead optimization
\ENDWHILE 

\RETURN $T_c$ 
\end{algorithmic} 
\end{algorithm}

\begin{table*}[t] 
\begin{center}
\renewcommand{\arraystretch}{1.2}
\begin{tabular}{|c|c|c|c|c|c|c|c|c|c|}
\hline

Dataset&Environment&Total Scans&deviation($t_x$)&deviation($t_y$)&deviation($t_z$)&deviation($\theta_x$)&deviation($\theta_y$) &deviation($\theta_z$)\\ [0.2ex]
\hline
\hline
$\begin{matrix} \text{Ford Campus} \\ \text{Vision and Lidar} \end{matrix}$ & Urban & 3538 & 0.65 & 2.87 & 0.05 & 0.02 & 0.02 & 0.23\\
\hline
$\begin{matrix} \text{Kitti Vision} \\ \text{Benchmark Suite} \end{matrix}$ & $\begin{matrix} \text{Rural /} \\ \text{Highway} \end{matrix}$ & 4919 & 2.67 & 0.31 & 0.04 & 0.01 & 0.01 & 0.08\\
\hline
\end{tabular}
\caption{Standard deviation (in meters for translation and radians for rotation) of the ground truth for all 6 DOF is shown. This analysis involves all the scan-pairs considered in our experiments. All floating point numeric values are rounded to two decimal places. It can be seen that the variance in $t_x$,$t_y$ and $\theta_z$ is more than the variance in $t_z$, $\theta_x$ and $\theta_y$. Although, we have shown this only for two different robots, this is in general true for any wheeled-robot. We use this information in initializing the simplex particles (N+1 points) in the optimization phase.} \label{table:gt_analysis}
\end{center}
\end{table*}

\subsection{Optimization}
The cost function (\ref{eq:cost}) is maximized at the correct value of rigid body transformation. Therefore, any optimization technique that iteratively converges to the local optimum can be used here. Some of the commonly used optimization techniques compute the gradient or hessian of the cost function (\cite{whittaker1967}, \cite{doi:10.1137/0111030} and \cite{10.2307/43633451}). There are also gradient free direct optimization methods like \textit{pattern search} \cite{doi:10.1137/S1052623493250780} or \textit{simulated annealing} \cite{Kirkpatrick671}. In this work we use one such direct optimization method called \textit{Nelder-Mead} optimization \cite{doi:10.1093/comjnl/7.4.308}.

Nelder-Mead method initializes a simplex with (N+1) points on the cost surface, where N is the number of DOF (six in our case) in the optimization phase. In each iteration it tries to improve these set of points by series of steps (reflecting, expanding, shrinking or contracting) to obtain the point which minimizes the cost function. NM-Simplex is sensitive to initial simplex size [$s_x$, $s_y$, $s_z$, $s_{roll}$, $s_{pitch}$, $s_{yaw}$]. 
Choosing an initial simplex large can cause unnecessary steps in areas of little interest, while a small simplex can lead to a narrow search on the cost surface increasing the computation. This problem can be alleviated by making a simple assumption that most of the relative traversal will be in $t_x$, $t_y$ and $\theta_z$ for wheeled robots. The intuition being that for any wheeled-robotic platform, motion is constrained in roll, pitch and z(height). Similar observation was also made in the localization method proposed in \cite{wolcott2015fast}. Making this assumption allows us to explore large range for $t_x$, $t_y$  and $\theta_z$ on cost surface, while at the same time a constrained initial search in $t_z$, $\theta_x$ and $\theta_y$. We verify this assumption, by calculating the deviation  of all 6-DOF in Table \ref{table:gt_analysis}. 

As this optimization technique is based on heuristics, optimization might not always lead to the optima. This is indeed a disadvantage of this optimization technique. While there exist methods that allow approximating the gradient of the cost function, however, the computation involved increases the run time of the algorithm. In the section below, we show that even after using a heuristic optimization, the results are close to the ground truth. 

The complete algorithm for obtaining the rigid body transformation is given in (Algorithm \ref{alg:alignment}).

\begin{figure*}[!ht]
\centering
\begin{subfigure}{\linewidth}
	\centering
    \includegraphics[width=\linewidth]{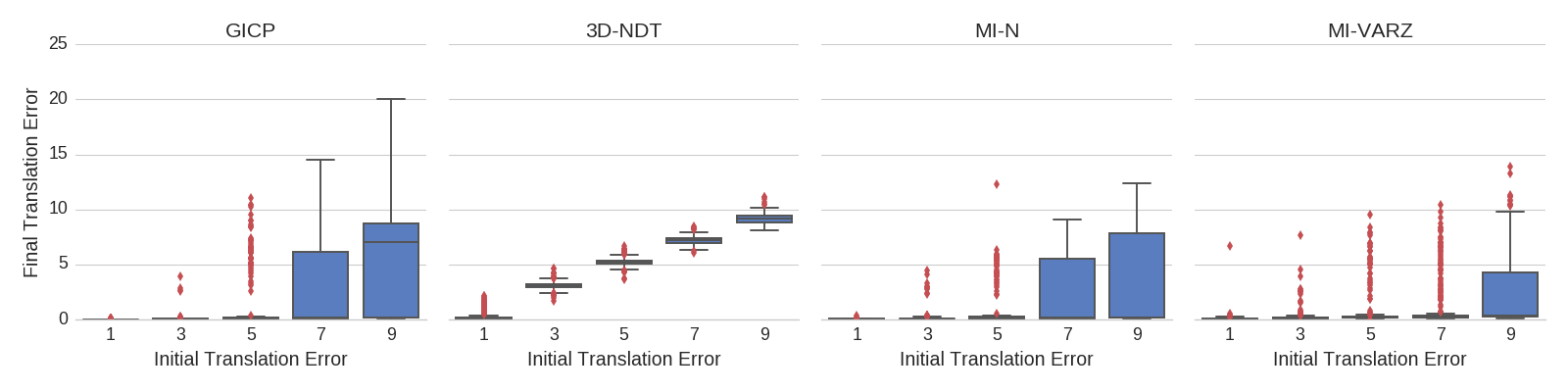}
    \caption{Boxplot depicting the variance and median of final translation error (in meters) versus initial error in translation parameters }\label{fig:ford_translation_error_plot}
\end{subfigure}

\begin{subfigure}{\linewidth}
    \centering
    \includegraphics[width=\linewidth]{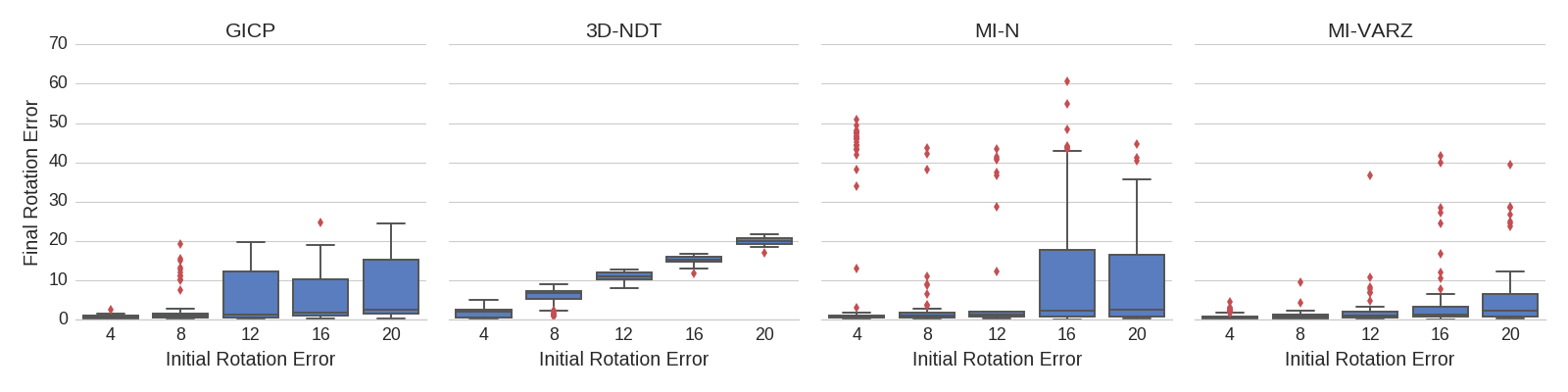}
    \caption{Boxplot depicting the variance and median of final rotation error (in degrees) versus initial error in rotation parameters }\label{fig:ford_rotation_error_plot}
\end{subfigure}

\begin{subfigure}{\linewidth}
	\centering
    \captionsetup{belowskip=10pt}
    \setlength\arrayrulewidth{0.7pt}
	\renewcommand{\arraystretch}{1.2}
	\begin{tabular}{|c|c|c|c|c|c|c|c|c|c|c|}
	\hline
	Method&1&2&3&4&5&6&7&8&9&10\\ [0.2ex]
	\hline
	GICP&\cellcolor{red!25}0.03&\cellcolor{red!25}0.07&\cellcolor{red!25}0.11&0.34&1.06&2.49&2.60&3.99&5.42&6.73\\
	3D-NDT&0.29&1.69&3.08&4.12&5.22&6.17&7.17&8.22&9.17&10.29\\
	MI-N&0.05&0.10&0.18&0.43&\cellcolor{red!25}0.68&1.22&2.09&2.68&2.80&3.23\\
	MI-VARZ&0.06&0.10&0.28&\cellcolor{red!25}0.30&0.70&\cellcolor{red!25}0.95&\cellcolor{red!25}1.37&\cellcolor{red!25}2.01&\cellcolor{red!25}2.39&\cellcolor{red!25}2.70\\
	\hline
	\end{tabular}
	\caption{Mean error (in meters) after alignment. Error in initial translation parameters is increased (from left: 1-10m).} \label{table:ford_mean_translation}
\end{subfigure}

\begin{subfigure}{\linewidth}
	\centering   
    \captionsetup{belowskip=10pt}
    \setlength\arrayrulewidth{0.7pt}
 	\renewcommand{\arraystretch}{1.2}
	\begin{tabular}{|c|c|c|c|c|c|c|c|c|c|c|}
	\hline
	Method&2&4&6&8&10&12&14&16&18&20\\ [0.2ex]
	\hline
GICP&\cellcolor{red!25}0.41&0.72&1.00&1.14&\cellcolor{red!25}0.88&\cellcolor{red!25}0.81&1.23&\cellcolor{red!25}1.78&3.26&5.05\\
    3D-NDT&0.52&1.82&4.48&5.98&8.96&11.03&13.17&15.24&17.43&19.44\\
	MI-N&1.18&1.90&2.39&4.79&4.19&5.66&2.60&5.61&4.91&1.49\\
	MI-VARZ&0.43&\cellcolor{red!25}0.70&\cellcolor{red!25}0.75&\cellcolor{red!25}0.72&0.92&2.56&\cellcolor{red!25}0.89&4.11&\cellcolor{red!25}1.98&\cellcolor{red!25}1.48\\
	\hline	    
	\end{tabular}   	
    \caption{Mean error (in degrees) after alignment. Error in initial rotation parameters is increased (from left: $2^\circ$-$20^\circ$).} \label{table:ford_mean_rotation}
\end{subfigure}

\caption{Ford Campus Vision and Lidar: Comparison between the proposed method (MI-VARZ and MI-N) with GICP and 3D-NDT. (a) and (b), depict the variance and median of the error magnitude (L2 norm) versus initial error for translation and rotation. In (c) and (d) we summarize the success rate of GICP, 3D-NDT, MI-N and MI-VARZ using mean error results in translation and rotation calculated by taking the L2 norm of difference with ground truth. Error analysis in (a) and (b) is done using descriptive statistical boxplots. In each boxplot, blue box represents the range from 25th percentile to 75th percentile and the black line inside the box shows the median. Results farther than 1.5 times the error at box-edges are considered outliers and are represented as red dots outside the whiskers. Ends of the whisker below and above the box represent the minimum and maximum values respectively.
} \label{ford_results}

\end{figure*}

\begin{figure*}[!ht]
\centering

\begin{subfigure}{\linewidth}
	\centering
    \includegraphics[width=\linewidth]{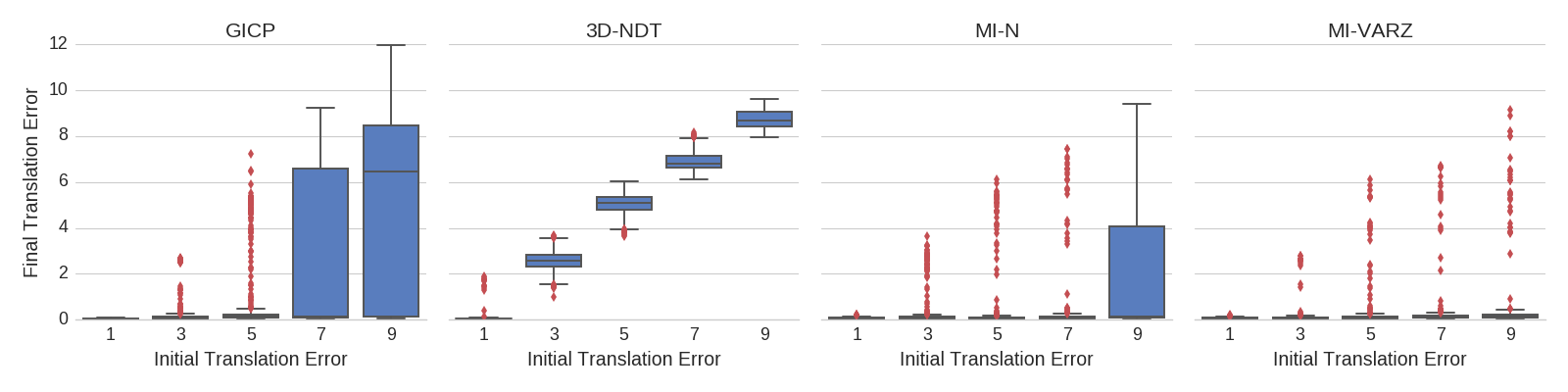}
    \caption{Boxplot depicting the variance and median of final translation error (in meters) versus initial error in translation parameters}\label{fig:kitti_translation_error_plot}
\end{subfigure}

\begin{subfigure}{\linewidth}
	\centering    
    \includegraphics[width=\linewidth]{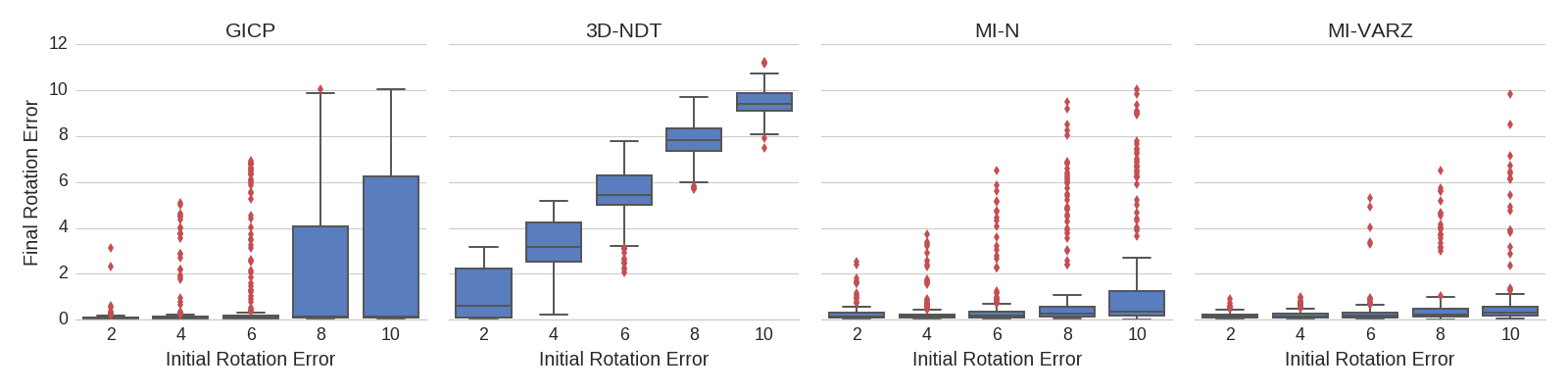}
    \caption{Boxplot depicting the variance and median of final rotation error (in degrees) versus initial error in rotation parameters}\label{fig:kitti_rotation_error_plot}
\end{subfigure}

\begin{subfigure}{\linewidth}
	\centering
    \captionsetup{belowskip=10pt}
    \setlength\arrayrulewidth{0.7pt}
	\renewcommand{\arraystretch}{1.2}
	\begin{tabular}{|c|c|c|c|c|c|c|c|c|c|c|c|}
	\hline
	Method&1&2&3&4&5&6&7&8&9&10\\ [0.2ex]
	\hline

GICP&\cellcolor{red!25}0.03&0.13&0.21&0.60&0.74&1.41&2.53&2.91&4.71&5.42\\
	3D-NDT&0.24&1.62&2.55&3.92&5.01&5.86&6.91&7.88&8.73&10.01\\
	MI-N&0.06&0.10&0.22&0.28&0.25&0.47&0.51&0.54&1.90&3.38\\
	MI-VARZ&0.06&\cellcolor{red!25}0.06&\cellcolor{red!25}0.12&\cellcolor{red!25}0.09&\cellcolor{red!25}0.21&\cellcolor{red!25}0.41&\cellcolor{red!25}0.42&\cellcolor{red!25}0.36&\cellcolor{red!25}0.68&\cellcolor{red!25}1.23\\
	\hline
	\end{tabular}
	\caption{Mean error (in meters) after alignment. Error in initial translation parameters in increased (from left: 1-10m).} \label{table:kitti_mean_translation}
\end{subfigure}

\begin{subfigure}{\linewidth}
	\centering
    \captionsetup{belowskip=10pt}
    \setlength\arrayrulewidth{0.7pt}
	\renewcommand{\arraystretch}{1.2}
	\begin{tabular}{|c|c|c|c|c|c|c|}
	\hline
	Method&2&4&6&8&10&12\\ [0.2ex]
	\hline

GICP&\cellcolor{red!25}0.13&0.34&0.82&2.08&2.70&2.90\\
	3D-NDT&1.10&3.21&5.43&7.82&9.49&11.47\\
	MI-N&0.30&0.27&0.51&1.20&1.85&\cellcolor{red!25}0.31\\
	MI-VARZ&0.17&\cellcolor{red!25}0.20&\cellcolor{red!25}0.31&\cellcolor{red!25}0.67&\cellcolor{red!25}0.94&0.36\\
	\hline
	\end{tabular}
	\caption{Mean error (in degrees) after alignment. Error in initial rotation parameters is increased (from left: $2^\circ$-$12^\circ$).} \label{table:kitti_mean_rotation}

\end{subfigure}

\caption{Kitti Vision and Benchmark Suite: Comparison between the proposed method (MI-VARZ and MI-N) with GICP and 3D-NDT. (a) and (b), depict the variance and median in the error for different magnitude error in initial estimate for translation and rotation. In (c) and (d) we summarize the success rate of GICP, 3D-NDT, MI-N and MI-VARZ using mean error results in translation and rotation calculated by taking the L2 norm of difference with ground truth. Error analysis in (a) and (b) is done using descriptive statistical boxplots. In each boxplot, blue box represents the range from 25th percentile to 75th percentile and the black line inside the box shows the median. Results farther than 1.5 times the error at box-edges are considered outliers and are represented as red dots outside the whiskers. Ends of the whisker below and above the box represent the minimum and maximum values respectively.
} \label{kitti_results}

\end{figure*}

\begin{figure*}[h]
\centering
\hspace*{-3mm}
\begin{subfigure}{.35\linewidth}
	\centering
    \includegraphics[width=\linewidth, height=\linewidth]{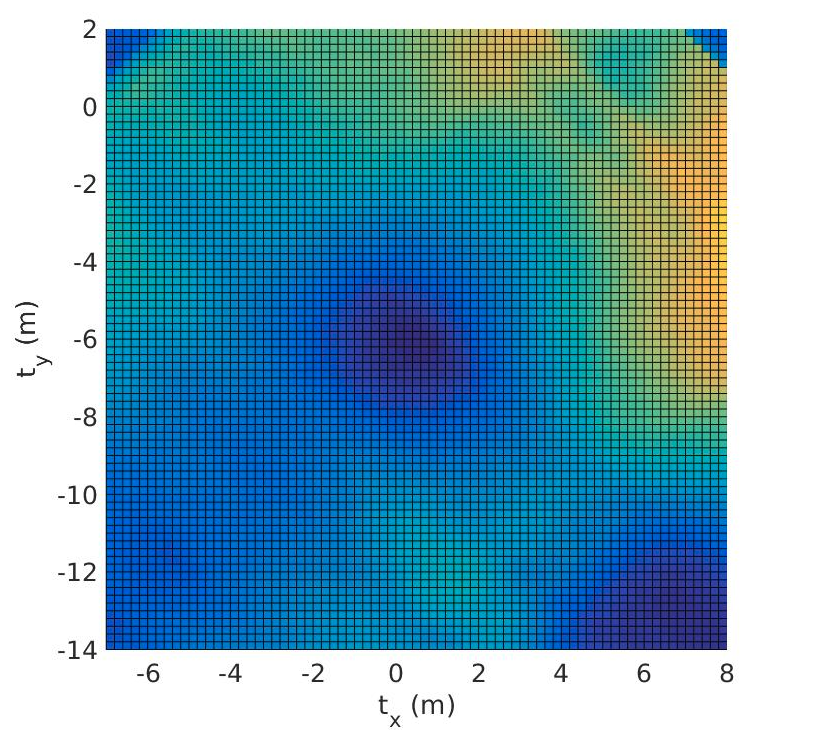}
    \caption{GICP cost versus $t_x$ and $t_y$}\label{fig:gicp_3D}
\end{subfigure}%
\begin{subfigure}{.35\linewidth}
    \centering
    \includegraphics[width=\linewidth, height=\linewidth]{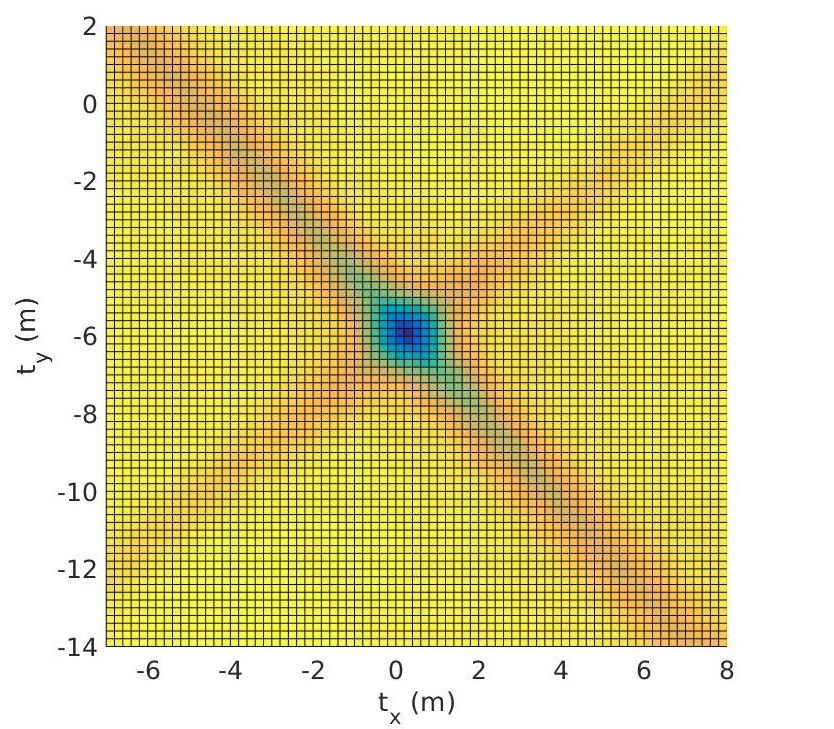}
    \caption{MI-VARZ cost versus $t_x$ and $t_y$}\label{fig:mi_3D}
\end{subfigure}%
\begin{subfigure}{.35\linewidth}
	\centering
    \includegraphics[width=\linewidth, height=\linewidth]{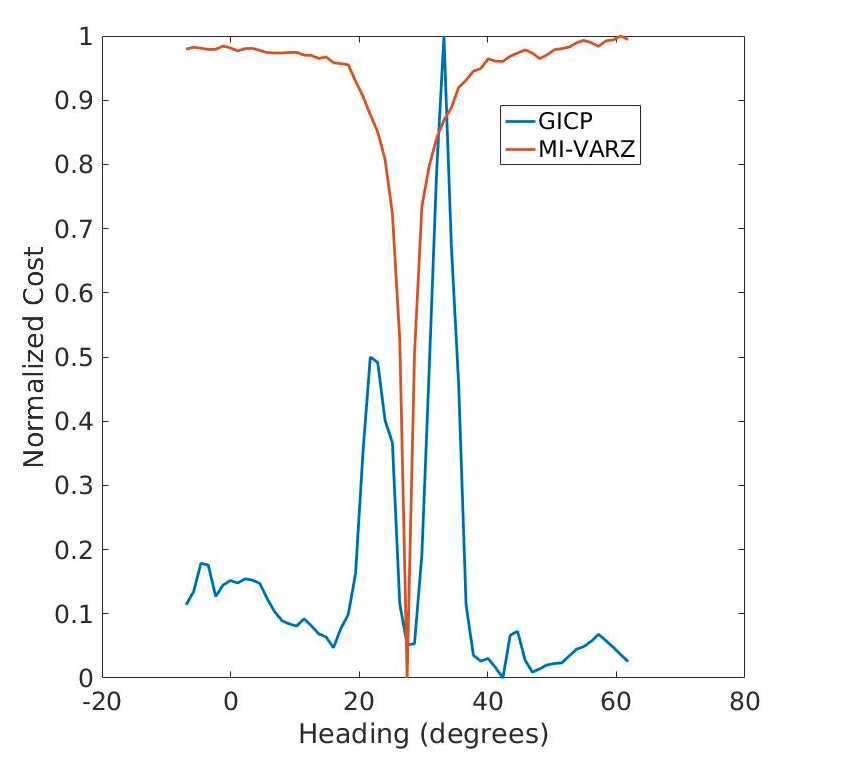}
    \caption{Normalized Cost versus heading}\label{fig:heading_cost}
\end{subfigure}
\caption{Cost-function plots for GICP and MI-VARZ. In (a) and (b), we have plotted the top view of GICP and MI-VARZ cost-function surface versus the translation parameters $t_x$ and $t_y$, while all other parameters are fixed to true ground-truth value. The correct translation parameters are $(t_x=0.40, t_y=-5.99)$. In (c), we compare the cost-function of GICP(Blue) and MI-VARZ(Orange) versus the heading. True heading parameter is $(\theta_z = 27.45^{\circ})$. It can be seen from both (a) and (c) that GICP can converge at a local optima far from global optimum when the initial error is large.} 
\label{fig:gicp_mi_comparison_plots}
\end{figure*}

\section{Experiments and Results} \label{evaluation}

We show results for two types of feature used in our MI framework: (\romannumeral 1) Variance of z-height in voxel (MI-VARZ), and (\romannumeral 2) Number of points in voxel (MI-N). We have used two point cloud datasets to assess the performance of the proposed approach. Scans from both these datasets are registered using GICP, 3D-NDT, MI-N, and MI-VARZ. The comparison factors were accuracy and runtime for different initial estimates. Datasets used in our experiments were Ford Campus Vision and Lidar dataset \cite{Pandey:2011:FCV:2049736.2049742} and odometry dataset from the Kitti Vision Benchmark Suite \cite{6248074}. Both the datasets consist of 3D scans collected from a test vehicle with a Lidar mounted on it and have ground truth pose information available from a highly accurate inertial navigation system (INS). 

Implementation of GICP is taken from the open source version available at \cite{gicp-link} and 3D-NDT is taken from Point Cloud Library (PCL) \cite{5980567}. Grid resolution ($R_v$) for 3D-NDT, MI-VARZ and MI-N is set as 1m. 
In our implementation we chose the initial simplex for optimization as [$s_x=8$, $s_y=8$, $s_z=1$, $s_{roll}=0.1$, $s_{pitch}=0.1$, $s_{yaw}=0.8$]. 

\subsection{Ford Campus Vision and Lidar Dataset}
In this experiment we compare the translation and rotation accuracy of the proposed method with GICP and 3D-NDT on Ford Dataset. To estimate the translation accuracy, we consider equi-spaced reference scan (A) in the downtown test run and sample 10 query scans (B) which are 1-10m from A. Similarly for rotation, all B query scans that have relative rotation upto $20^\circ$ (with translation error less than 5m) from scan A are chosen. In total 3538 scan pairs from Ford dataset were considered to plot error results shown in Figure \ref{ford_results}. 

\subsection{Kitti Vision Benchmark Suite}
In this experiment, we use the scans from Kitti dataset for testing both translation and rotation accuracy. Here also, we sample scans that have translation 1-10m and rotation upto $12^\circ$ (plot range is reduced here as the number of scan pairs with translation error less than 5m and rotation greater than $12^\circ$ are not enough) resulting in a total of 4919 scan pairs from a single session run. Results are shown in Figure \ref{kitti_results}.

\subsection{Discussion of results}
It can be seen, that when the initial estimate of transformation parameters are good, all 4 methods perform reasonably good for both the datasets. However, 3D-NDT fails to converge when the initial error is increased: this can be explained by the discontinuity in score function due to the jump caused when the query scan $B$ passes one of the cell boundaries. In \cite{5152538} and \cite{4058864}, convergence of 3D-NDT and it's dependence on voxels resolution is explained in detail. 

It should be noted that GICP works better when we have a good initial guess for translation parameters. However, it fails to converge to the correct optima when the initial guess is poor. This can be explained from cost surface of GICP algorithm. As seen in Figure \ref{fig:gicp_3D}, GICP has several local optima causing the gradient based optimization methods to fail. Figure \ref{fig:heading_cost} also depicts this problem, when the initial heading error is large GICP can converge to incorrect transformation. In contrast, in Figure \ref{fig:mi_3D} it can be seen that our MI based method has a single optima corresponding to the correct rigid-body-transformation for the same range of initial error. It is important to note that this might not always be the case. In scan pairs with multiple dynamic objects and large initial error, the heuristic based optimization in MI-VARZ and MI-N can converge at an optima far from the correct solution. Despite that, we observe that our MI based method performs better than GICP in most cases.

Both MI-VARZ and MI-N have better results for translation as we increase the prior translation error. For the Kitti Dataset, it can be seen that MI-VARZ has average translation error less than 0.5m for the initial translation error up till 8m and MI-N till 6m, which is in contrast to results from GICP where we see an average error of 0.6m at 4m of initial translation error. In the Ford dataset, MI-VARZ and MI-N have error less than 0.5m up till 4m and GICP till 3m. 

As for the rotation error, in Kitti Dataset we see a similar trend where MI-VARZ performs better than GICP as we increase the initial error. In the Ford dataset however, both GICP and MI-VARZ produce mixed results. This is primarily due to unstructured areas and multiple dynamic objects in the urban scene. In such cases, the heuristic based search in MI-VARZ leads to outliers (See Figure \ref{fig:ford_rotation_error_plot} and \ref{table:ford_mean_rotation} for initial error $16^{\circ}$) which are relatively farther from the correct rotation parameters than GICP outliers. Yet in most cases (Figure \ref{fig:ford_rotation_error_plot}) MI-VARZ has performance better than GICP. We also observe that MI-N doesn't perform as good as MI-VARZ and GICP. In general too, MI-N has less success rate as compared to MI-VARZ, this is because the number of points in a voxel do not convey any information about the local-structure unlike the case in variance of z.

\subsection{Runtime analysis}


In this section we compare the runtime of GICP with MI-N and MI-VARZ. We do not consider the 3D-NDT in this analysis as it fails to converge when the initial error is high. The average runtime of GICP, MI-N and MI-VARZ is shown in Table \ref{table:runtime}. All the implementations were executed on a system powered by Intel Core i7-7700HQ CPU@ 2.80GHz $\times$ 8 and GeForce GTX 1050Ti.	We used \cite{muja_flann_2009} to optimize the GICP by implementing the point-point search queries in GPU memory. The runtime of GICP after this optimization is also showed in Table \ref{table:runtime}. 

\begin{table}[h] 
\begin{center}
\renewcommand{\arraystretch}{1.5}
\begin{tabular}{|c c c c c|}
\hline
Initial Error & GICP & GICP-GPU & MI-N & MI-VARZ \\ [0.5ex]
(m) & (s) & (s) & (s) & (s)\\ [0.5ex]
\hline
\hline
1 & 2.67 & 1.64 & 1.26 & 0.49 \\ 
3 & 4.00 & 2.25 & 1.27 & 0.50 \\ 
5 & 6.92 & 4.42 & 1.26 & 0.50 \\ 
7 & 10.72 & 7.25 & 1.23 & 0.49 \\ 
9 & 14.26 & 10.13 & 1.22 & 0.51 \\ 
\hline
\end{tabular}
\caption{Runtime analysis of GICP (Original), GICP (GPU Search), MI-N and MI-VARZ. MI-N and MI-VARZ have negligible variability in runtime with respect to quality of the initial guess. However, the GICP runtime increases as the initial error is increased.}
\label{table:runtime}
\end{center}
\end{table}

Another important challenge is the variability of runtime with change in the initial error. The runtime of GICP varies when we change the error in translation, whereas MI-N and MI-VARZ have runtime independent of the initial error. 

In Table \ref{table:gpu_operations}, we show that all major steps in our proposed framework can be easily offloaded to GPU with minimal overhead. This is in contrast to other point-to-point alignment methods like GICP, which require building nearest neighbor structures (like K-D Tree) in GPU. The preprocessing step in GICP for nearest neighbor search takes most of the time in alignment process. Whereas in our MI based framework, the only bottleneck is GPU memory allocation (along with gpu context initialization).

\begin{table}[!h] 
\begin{center}
\renewcommand{\arraystretch}{1.2}
\begin{tabular}{|ll|c|c|}
\hline
MI Steps&Kernel&Total Time(ms)\\ [0.2ex]
\hline
\hline
GPU Allocate memory &-&189.00\\
Data CPU to GPU&-&0.5\\
Voxel Mapping&Map&5.87\\
Find overlap&Reduce&33.06\\
Feature histograms&Map&100.10\\
Entropy Calculation& Reduce&8.45\\
\hline
\end{tabular}
\caption{GPU profiling for a scan pair while executing MI-VARZ. Second column depicts the type of gpu kernel for the operation and third column is the total time (ms) spent in kernel execution for all iterations in the alignment process.}
\label{table:gpu_operations}
\end{center}
\end{table}

\section{CONCLUSION}
In this paper, we report a MI based scan alignment algorithm that maximizes the mutual information between the voxelized features of two partially overlapping scans by calculating a single-dimensional feature in a voxel. Our approach allows us to consider the no-feature voxels in our cost function with the intuition being: that both scans, when aligned, should have same number of unoccupied voxels in the overlapping region. The proposed method is tested and compared with a point-to-point and point-to-distribution method on two real-world datasets covering wide range of dynamic scenes: urban and rural. We see that our method performs relatively better even for large initial errors in transformation. Although, we implemented our method with data from a single sensor (lidar), it can easily be extended to fused-sensor data (lidar-camera). We show that our method has a fast GPU implementation which allows computation to be offloaded to GPU with minimal overhead.


\bibliographystyle{ieeetr}
\bibliography{ref}

\begin{thebibliography}{10}

\bibitem{Besl:1992:MRS:132013.132022}
P.~J. Besl and N.~D. McKay, ``A method for registration of 3-d shapes,'' {\em
  IEEE Trans. Pattern Anal. Mach. Intell.}, vol.~14, pp.~239--256, Feb. 1992.

\bibitem{1240280}
M.~Greenspan and M.~Yurick, ``Approximate k-d tree search for efficient icp,''
  in {\em Fourth International Conference on 3-D Digital Imaging and Modeling,
  2003. 3DIM 2003. Proceedings.}, pp.~442--448, Oct 2003.

\bibitem{Lu1997}
F.~Lu and E.~Milios, ``Robot pose estimation in unknown environments by
  matching 2d range scans,'' {\em Journal of Intelligent and Robotic Systems},
  vol.~18, no.~3, pp.~249--275, 1997.

\bibitem{55667789-MIT}
M.~C. Bosse, {\em ATLAS: a framework for large scale automated mapping and
  localization}.
\newblock PhD thesis, Massachusetts Institute of Technology. Dept. of
  Electrical Engineering and Computer Science, 2004.

\bibitem{297537764-MIT}
E.~B. E.~B. Olson, {\em Robust and efficient robotic mapping}.
\newblock PhD thesis, Massachusetts Institute of Technology. Dept. of
  Electrical Engineering and Computer Science, 2008.

\bibitem{Censi08anicp}
A.~Censi, ``An icp variant using a point-to-line metric,'' in {\em IEEE
  International Conference on Robotics and Automation}, 2008.

\bibitem{Segal-RSS-09}
A.~Segal, D.~Haehnel, and S.~Thrun, ``Generalized-icp,'' in {\em Proceedings of
  Robotics: Science and Systems}, (Seattle, USA), June 2009.

\bibitem{Biber2003TheND}
P.~Biber and W.~Stra{\ss}er, ``The normal distributions transform: a new
  approach to laser scan matching,'' in {\em IROS}, 2003.

\bibitem{Magnusson053dmodelling}
M.~Magnusson, R.~Elsrud, L.~erik Skagerlund, and T.~Duckett, ``3d modelling for
  underground mining vehicles,'' in {\em In SimSafe 2005, Proceedings of the
  Conference on Modeling and Simulation for Public Safety}, 2005.

\bibitem{kim2016}
J.~W. Kim and B.~H. Lee, ``Robust and fast 3-d scan registration using normal
  distributions transform with supervoxel segmentation,'' {\em Robotica},
  vol.~34, pp.~1630--1658, 007 2016.

\bibitem{Johnson:1997:RIT:523428.825385}
A.~E. Johnson and S.~B. Kang, ``Registration and integration of textured 3-d
  data,'' in {\em Proceedings of the International Conference on Recent
  Advances in 3-D Digital Imaging and Modeling}, NRC '97, (Washington, DC,
  USA), pp.~234--, IEEE Computer Society, 1997.

\bibitem{Godin2001AMF}
G.~Godin, D.~Laurendeau, and R.~Bergevin, ``A method for the registration of
  attributed range images,'' in {\em 3DIM}, 2001.

\bibitem{Lowe:2004:DIF:993451.996342}
D.~G. Lowe, ``Distinctive image features from scale-invariant keypoints,'' {\em
  Int. J. Comput. Vision}, vol.~60, pp.~91--110, Nov. 2004.

\bibitem{f8a4caebc7cf412ca3ed2113bf6c94df}
G.~Pandey, J.~McBride, S.~Savarese, and R.~Eustice, {\em Visually bootstrapped
  generalized ICP}, pp.~2660--2667.
\newblock 2011.

\bibitem{pmid8331222}
R.~P. Woods, J.~C. Mazziotta, and S.~R. Cherry, ``{{M}{R}{I}-{P}{E}{T}
  registration with automated algorithm},'' {\em J Comput Assist Tomogr},
  vol.~17, no.~4, pp.~536--546, 1993.

\bibitem{Hill1993}
D.~L.~G. Hill, D.~J. Hawkes, N.~A. Harrison, and C.~F. Ruff, {\em A strategy
  for automated multimodality image registration incorporating anatomical
  knowledge and imager characteristics}, pp.~182--196.
\newblock Berlin, Heidelberg: Springer Berlin Heidelberg, 1993.

\bibitem{Viola1997}
P.~Viola and W.~M. Wells~III, ``Alignment by maximization of mutual
  information,'' {\em International Journal of Computer Vision}, vol.~24,
  no.~2, pp.~137--154, 1997.

\bibitem{pmid9101328}
F.~Maes, A.~Collignon, D.~Vandermeulen, G.~Marchal, and P.~Suetens,
  ``{{M}ultimodality image registration by maximization of mutual
  information},'' {\em IEEE Trans Med Imaging}, vol.~16, pp.~187--198, Apr
  1997.

\bibitem{pmid10709702}
F.~Maes, D.~Vandermeulen, and P.~Suetens, ``{{C}omparative evaluation of
  multiresolution optimization strategies for multimodality image registration
  by maximization of mutual information},'' {\em Med Image Anal}, vol.~3,
  pp.~373--386, Dec 1999.

\bibitem{6386053}
G.~Pandey, J.~R. McBride, S.~Savarese, and R.~M. Eustice, ``Toward mutual
  information based automatic registration of 3d point clouds,'' in {\em 2012
  IEEE/RSJ International Conference on Intelligent Robots and Systems},
  pp.~2698--2704, Oct 2012.

\bibitem{1216223}
J.~P.~W. Pluim, J.~B.~A. Maintz, and M.~A. Viergever,
  ``Mutual-information-based registration of medical images: a survey,'' {\em
  IEEE Transactions on Medical Imaging}, vol.~22, pp.~986--1004, Aug 2003.

\bibitem{whittaker1967}
E.~T. Whittaker and G.~Robinson, ``The newton raphson method,'' in {\em The
  Calculus of Observations: A Treatise on Numerical Mathematics}, vol.~44,
  pp.~84--87, 1967.

\bibitem{doi:10.1137/0111030}
D.~W. Marquardt, ``An algorithm for least-squares estimation of nonlinear
  parameters,'' {\em Journal of the Society for Industrial and Applied
  Mathematics}, vol.~11, no.~2, pp.~431--441, 1963.

\bibitem{10.2307/43633451}
K.~LEVENBERG, ``A method for the solution of certain non-linear problems in
  least squares,'' {\em Quarterly of Applied Mathematics}, vol.~2, no.~2,
  pp.~164--168, 1944.

\bibitem{doi:10.1137/S1052623493250780}
V.~Torczon, ``On the convergence of pattern search algorithms,'' {\em SIAM
  Journal on Optimization}, vol.~7, no.~1, pp.~1--25, 1997.

\bibitem{Kirkpatrick671}
S.~Kirkpatrick, C.~D. Gelatt, and M.~P. Vecchi, ``Optimization by simulated
  annealing,'' {\em Science}, vol.~220, no.~4598, pp.~671--680, 1983.

\bibitem{doi:10.1093/comjnl/7.4.308}
J.~A. Nelder and R.~Mead, ``A simplex method for function minimization,'' {\em
  The Computer Journal}, vol.~7, no.~4, p.~308, 1965.

\bibitem{wolcott2015fast}
R.~W. Wolcott and R.~M. Eustice, ``Fast lidar localization using
  multiresolution gaussian mixture maps,'' in {\em Robotics and Automation
  (ICRA), 2015 IEEE International Conference on}, pp.~2814--2821, IEEE, 2015.

\bibitem{Pandey:2011:FCV:2049736.2049742}
G.~Pandey, J.~R. Mcbride, and R.~M. Eustice, ``Ford campus vision and lidar
  data set,'' {\em Int. J. Rob. Res.}, vol.~30, pp.~1543--1552, Nov. 2011.

\bibitem{6248074}
A.~Geiger, P.~Lenz, and R.~Urtasun, ``Are we ready for autonomous driving? the
  kitti vision benchmark suite,'' in {\em 2012 IEEE Conference on Computer
  Vision and Pattern Recognition}, pp.~3354--3361, June 2012.

\bibitem{gicp-link}
A.~Segal, ``Generalized icp reference implementation.''

\bibitem{5980567}
R.~B. Rusu and S.~Cousins, ``3d is here: Point cloud library (pcl),'' in {\em
  2011 IEEE International Conference on Robotics and Automation}, pp.~1--4, May
  2011.

\bibitem{5152538}
M.~Magnusson, A.~Nuchter, C.~Lorken, A.~J. Lilienthal, and J.~Hertzberg,
  ``Evaluation of 3d registration reliability and speed - a comparison of icp
  and ndt,'' in {\em 2009 IEEE International Conference on Robotics and
  Automation}, pp.~3907--3912, May 2009.

\bibitem{4058864}
E.~Takeuchi and T.~Tsubouchi, ``A 3-d scan matching using improved 3-d normal
  distributions transform for mobile robotic mapping,'' in {\em 2006 IEEE/RSJ
  International Conference on Intelligent Robots and Systems}, pp.~3068--3073,
  Oct 2006.

\bibitem{muja_flann_2009}
M.~Muja and D.~G. Lowe, ``Fast approximate nearest neighbors with automatic
  algorithm configuration,'' in {\em International Conference on Computer
  Vision Theory and Application VISSAPP'09)}, pp.~331--340, INSTICC Press,
  2009.

\end{thebibliography}

\end{document}